\documentclass[letterpaper, 10 pt, conference]{ieeeconf}  

\IEEEoverridecommandlockouts                              

\usepackage{graphicx} 
\usepackage{amsmath} 
\usepackage{amssymb}  
\usepackage[ruled,linesnumbered]{algorithm2e}
\usepackage{bm}
\usepackage{booktabs}
\usepackage{threeparttable}
\usepackage{cite}
\usepackage{multirow}
\usepackage{multicol}
\usepackage{backref}
\usepackage{url}
\usepackage{hyperref}
\usepackage{balance}
\usepackage{algpseudocode}

\hypersetup{
    colorlinks=true, 
    urlcolor=magenta,
}

\title{\LARGE \bf
HAPI: A Model for Learning Robot Facial Expressions from Human Preferences
}

\author{Dongsheng Yang$^{1,2}$, Qianying Liu$^{5}$, Wataru Sato$^{2,3}$, Takashi Minato$^{4}$, Chaoran Liu$^{5}$, Shin'ya Nishida$^{1}$
\thanks{$^{1}$Graduate School of Informatics, Kyoto University, Kyoto, Japan
    {\tt\small mlds.yang@gmail.com}, $^{2}$Psychological Process Research Team, Guardian Robot Project, RIKEN, Kyoto, Japan, $^{3}$Field Science Education and Research Center, Kyoto University, Kyoto, Japan, $^{4}$Interactive Robot Research Team, Guardian Robot Project, RIKEN, Kyoto, Japan and $^{5}$Research and Development Center for Large Language Models, National Institute of Informatics, Tokyo, Japan}%
    \thanks{This work is supported by JSP.}
}

\begin{document}

\maketitle
\thispagestyle{empty}
\pagestyle{empty}

\begin{abstract}

Automatic robotic facial expression generation is crucial for human–robot interaction (HRI), as handcrafted methods based on fixed joint configurations often yield rigid and unnatural behaviors. Although recent automated techniques reduce the need for manual tuning, they tend to fall short by not adequately bridging the gap between human preferences and model predictions—resulting in a deficiency of nuanced and realistic expressions due to limited degrees of freedom and insufficient perceptual integration. In this work, we propose a novel learning-to-rank framework that leverages human feedback to address this discrepancy and enhanced the expressiveness of robotic faces. Specifically, we conduct pairwise comparison annotations to collect human preference data and develop the Human Affective Pairwise Impressions (HAPI) model, a Siamese RankNet-based approach that refines expression evaluation. Results obtained via Bayesian Optimization and online expression survey on a 35-DOF android platform demonstrate that our approach produces significantly more realistic and socially resonant expressions of Anger, Happiness, and Surprise than those generated by baseline and expert-designed methods. This confirms that our framework effectively bridges the gap between human preferences and model predictions while robustly aligning robotic expression generation with human affective responses.

\end{abstract}


\section{INTRODUCTION}

Facial expressions are a fundamental channel of nonverbal communication, conveying up to 55\% of interpersonal cues~\cite{mehrabian2017communication}, and are pivotal for effective HRI. Realistic and socially resonant robotic facial expressions (RFEs) facilitate nuanced and natural interaction by enhancing both emotional conveyance and intention recognition. 

Despite the important role of RFEs for effective HRI, most existing techniques rely on preprogrammed joint configurations, mapping each RFE to a fixed set of mechanical movements~\cite{breazeal2003emotion,kim2006development,bennett2014deriving,silva2016mirroring}. Although these rule-based methods could explore a range of RFEs, they typically produce rigid, unnatural behaviors and required extensive manual effort, including hardware-specific adaptations, while offering limited generalization capabilities.

\begin{figure}[t]
  \centering
  \includegraphics[width=0.45\textwidth]{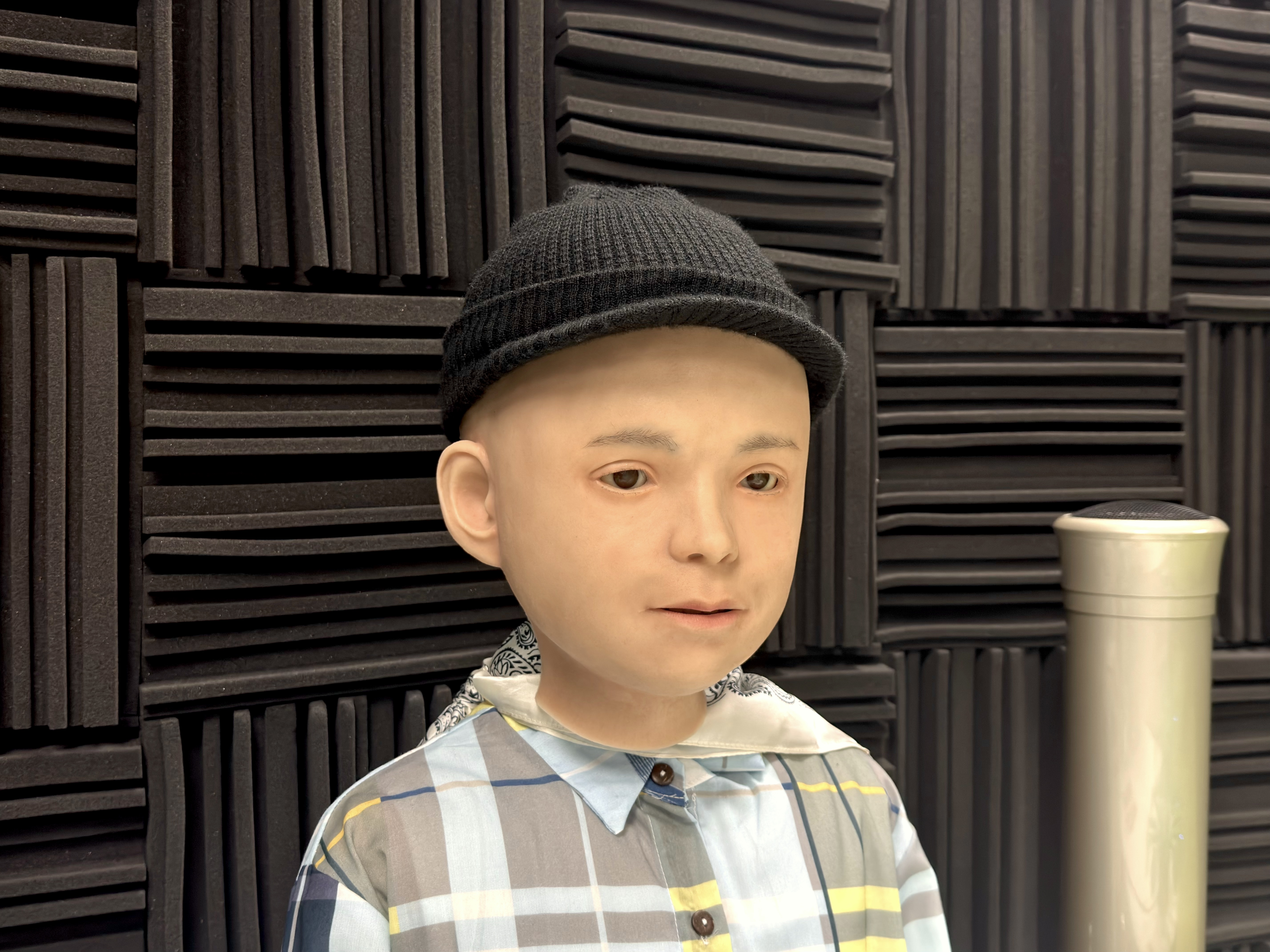}
  \caption{Nikola, an android developed by RIKEN.}
  \label{fig:intro}
\end{figure}

Recent advances have sought to automate RFE generation. Early work~\cite{breazeal2005learning,ren2016automatic,chen2021smile} introduced algorithms that reduced the need for hard-coding. Yang et al.~\cite{yang2022BORFEO} proposed a Bayesian Optimization-based framework for automatic RFE generation that incorporated both mechanical stress and actuator non-linearity considerations. Their method, implemented on Nikola (Figure \ref{fig:intro}), a cutting-edge 35-DOF android developed by RIKEN, relied on a vision-based facial expression recognition model (ResMaskNet~\cite{pham2021facial}) rather than marker-based sensing. Although this approach~\cite{yang2022BORFEO} achieved mechanical efficiency and reduced trial requirements compared to other learning-based methods, it struggled to generate convincing expressions of happiness—even though happiness is one of the most universally recognized emotions.  

We observe that RFEs with the highest ResMaskNet ratings for happiness received low human evaluations, highlighting a misalignment between ResMaskNet's quantitative metrics and human perceptual judgment~\cite{yang2022BORFEO}. This limitation arises because ResMaskNet's training objective was categorizing human facial expressions, which neglected the subtle nuances essential for conveying complex emotions—a criterion fundamentally distinct from the aims of generating realistic and socially resonant RFEs. 

\begin{figure*}[t]
  \centering
  \includegraphics[width=\textwidth]{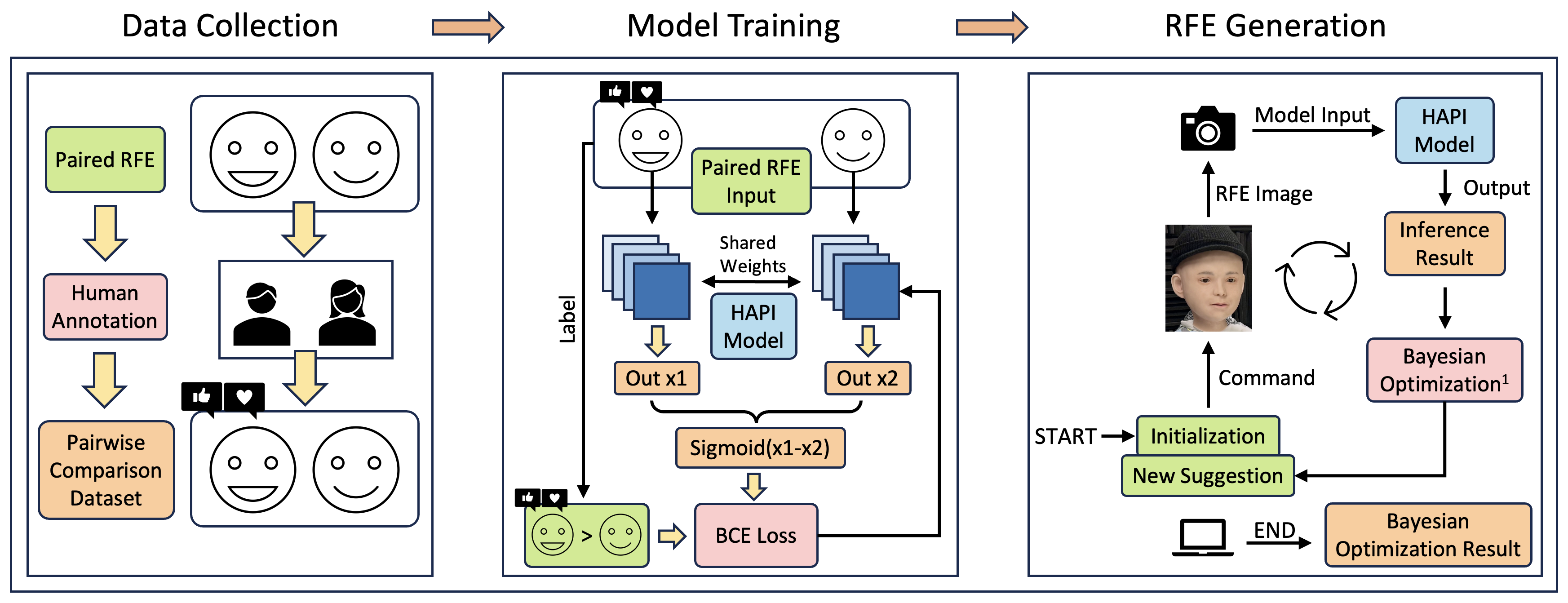}
  \caption{Overview of our three‐stage pipeline for robotic facial expression (RFE) generation. \textbf{(Left)} We collect a pairwise comparison dataset by asking human evaluators to choose preferred expressions from paired RFEs. \textbf{(Center)} The Human Affective Pairwise Impressions (HAPI) model—a Siamese RankNet—learns from these pairwise preferences, refining its ability to discern subtle perceptual nuances. \textbf{(Right)} Using the trained HAPI model, we employ Bayesian optimization to iteratively suggest and evaluate new RFEs on the 35‐DOF android robot, ultimately converging on expressions that align closely with human affective judgments.}
  \label{fig:pipe}
\end{figure*}

To address the limitation that existing models generate only expression probabilities without incorporating human perceptual insights, we aim to leverage human feedback to enhance the realism and social resonance of robotic facial expressions. However, modeling subtle RFEs is inherently complex and demands extensive human labeling, which often leads to data scarcity. To tackle these issues, this study employs a pairwise comparison annotation method to collect a human preference dataset on robotic facial expressions, leveraging pairwise comparison as an effective psychophysical procedure for assessing perceptual preferences, enlightened by Thurstone~\cite{thurstone1927law}. Building on this dataset, we design the model structure inspired by reinforcement learning from human feedback (RLHF)~\cite{christiano2017deep, ouyang2022training}, leveraging human preferences to fine-tune robotic expressions to enhance realism and social resonance. While RLHF has mainly been applied to robotic manipulation tasks~\cite{hiranaka2023primitive, balsells2023autonomous}, to the best of our knowledge, 
this work represents the first attempt to draw on human feedback-inspired approaches for facial expression generation on an android robot. 

By leveraging subtle and subjective human preference data from robot-generated emotional images, we propose a novel approach termed Human Affective Pairwise Impressions (HAPI). HAPI employs a Siamese RankNet architecture to refine the evaluation process, ensuring that robotic expressions progressively align with nuanced human preferences while capturing the intricacies of human affective perception. 
This approach constructs a learning-to-rank framework that utilizes pairwise comparison annotations to establish a more perceptually grounded evaluation metric, allowing robotic expressions to be assessed and ranked based on human affective judgments.
Consequently, HAPI enhances both the realism and social resonance of robotic facial expressions while improving sensitivity to subtle variations in human affect, ultimately leading to more robust and contextually attuned expressions.

In summary, we propose a novel learning-to-rank framework that models human preferences in evaluating robotic facial expressions. Furthermore, we apply this framework to guide the generation of robotic facial expressions, enhancing their quality. An overview of our approach is illustrated in Figure~\ref{fig:pipe}. 

Our contributions are as follows:
\begin{enumerate} 
    \item We conduct a pairwise comparison annotation to collect a human preference dataset for robot-generated facial expressions. 
    \item We introduce HAPI, a learning framework that leverages subtle and subjective human preference data through a Siamese RankNet architecture to enhance the realism and expressiveness of robotic facial behaviors, ensuring that the generated expressions progressively align with nuanced human expectations. 
    \item We verify the effectiveness of our approach by applying Bayesian Optimization and conducting online human survey experiments. Our results demonstrate that the robot facial expressions generated by our model outperform both baseline methods and expert-designed prototypes in Anger, Happiness, and Surprise expressions.
\end{enumerate}

\section{RELATED WORK}

Automated systems for recognizing facial expressions have long been a focus of research \cite{li2022deepfersurvey}, yet the generation of facial expressions on android robots for human-robot interaction has only recently garnered significant attention \cite{rawal2022exgennet, chen2021smile}. Traditionally, many approaches depended on predefined joint configurations, where servo motor movements—such as those for the eyelids and mouth—were manually programmed to represent basic expressions \cite{breazeal2003emotion, venture2019}. More sophisticated methods have employed the Facial Action Coding System (FACS)~\cite{ekman1978facial} to capture subtle facial cues, allowing human experts to fine-tune expressions based on nuanced FACS signals.

A range of studies has pursued automated facial expression generation. Early work by Breazeal et al. \cite{breazeal2005learning} demonstrated the use of simple neural networks to produce basic expressions on the humanoid robot Leonardo. Building on this, Horii et al. \cite{horii2016human} utilized Restricted Boltzmann Machines (RBM) to model and synthesize facial expressions for emotion-based imitation tasks, while Ren et al. \cite{ren2016automatic} explored kinematics-based learning for expression generation on the XIN-REN android robot. More recently, Churamani et al. \cite{churamani2018learning} applied reinforcement learning to optimize expressions through iterative feedback, highlighting the potential of adaptive methods in this domain.

Further advancements included the vision-based self-supervised learning framework developed by Chen et al. \cite{chen2021smile}, which enabled robots to mimic human facial expressions without requiring prior knowledge of the robot's kinematics. In parallel, Hu et al. \cite{hu2024human} introduced a method that allowed robots to predict and coexpressed human smiles by learning an inverse kinematic facial self-model, and Yuan et al. \cite{yuan2024two} proposed a two-stage facial kinematic control strategy—utilizing keyframe detection and cubic spline interpolation—to achieve natural and smooth transitions in facial expressions. Additionally, Wu et al. \cite{wu2024retargeting} presented an innovative approach that leveraged neural network surrogate models to retarget human expressions onto robotic faces, further enhancing the expressiveness and naturalness of robotic facial behaviors.

Notably, Bayesian Optimization has emerged as a promising approach to refine robotic facial expressions. Yang et al. \cite{yang2022BORFEO} demonstrated that a Bayesian framework could significantly reduce the search time for achieving appropriate expressions while accommodating mechanical constraints inherent to robotic platforms. However, challenges remained in aligning the generated expressions with human affective preferences. Misalignment in the emotion model can result in expressions that are perceived as inauthentic or inappropriate, potentially undermining the effectiveness of human-robot interactions.

Collectively, these studies illustrate the evolution from manually programmed expressions to advanced, automated systems that integrate machine learning and optimization techniques. This progression represents an ongoing effort to bridge the gap between the mechanical limitations of robotic platforms and the rich, nuanced dynamics of human facial expressions, thereby enhancing the overall quality of human-robot interaction.

\section{METHOD}

We define our task in a framework \textit{learning from human feedback}, where we lack a direct, absolute measure of the intensity of robot facial expression. Therefore, we collect binary comparison labels \(I_{i} \succ I_{j}\) from human annotators, indicating that expression \(I_{i}\) is considered more (or better) than \(I_{j}\). This approach aligns with previous work on pairwise comparisons for human feedback \cite{Yan_2014_CVPR,bai2022training,Furnkranz2003} and naturally lends itself to a \textit{learning-to-rank} paradigm \cite{Furnkranz2003}. Specifically, our method is divided into three phases: (1) constructing a human preference dataset using pairwise annotations, (2) fine-tuning a robot expression recognition model on those annotated data, 
and (3) deploying the trained model to evaluate and refine robot-generated facial expressions, benchmarking its performance against two baseline methods in a controlled experimental setup.

\subsection{Human Preference Dataset on Robot Facial Expressions}\label{sec:hpd}

To construct a preference-based evaluation framework for robotic facial expressions, we first generate a set of RFEs by leveraging a Bayesian Optimization framework on generating RFEs\cite{yang2022BORFEO}. Specifically, instead of randomly sampling from the full parameter space— which may result in ineffective expressions—we collect the optimized samples discovered during the Bayesian optimization process. These expressions have a high probability of closely approximating the target expression, ensuring their relevance and validity. Additionally, due to the inherent exploration mechanism of Bayesian optimization, the collected samples exhibit a certain degree of diversity, further enriching the dataset.

Then, to ensure a diverse subset of robotic facial expressions, we select 100 images from a larger candidate pool by maximizing expression variability using pixel-wise cosine distance \( S_C \) (Eq.~\ref{eq:cosine}).
From an initial pool of 100 RFE images, we form \(\binom{100}{2} = 4950\) distinct pairs for evaluation. An online platform then collect pairwise comparisons indicating which image in each pair is preferred.

\begin{equation}
{\displaystyle {S_{C}(A,B):={\mathbf {A} \cdot \mathbf {B}  \over \|\mathbf {A} \|\|\mathbf {B} \|}={\frac {\sum \limits _{i=1}^{n}{A_{i}B_{i}}}{{\sqrt {\sum \limits _{i=1}^{n}{A_{i}^{2}}}}\cdot {\sqrt {\sum \limits _{i=1}^{n}{B_{i}^{2}}}}}}}}
\label{eq:cosine}
\end{equation}

We note that while exhaustive pairwise annotation becomes infeasible as the dataset size increases, efficient insertion strategies such as binary search can address this challenge by determining the position of a new image within the existing ranking using only \(O(\log n)\) additional comparisons, rather than requiring a complete set of pairwise evaluations.

Formally, we construct a ranking to model the observed pairwise preferences among images. Each image \(I_i\) is associated with an unknown true score \(f^*(I_i)\); however, rather than accessing these scores directly, we rely on the relative comparisons \(I_i \succ I_j\) to indicate that image \(I_i\) is preferred over \(I_j\). While completing the permutation of the dataset \(D = \{I_1, I_2, \dots, I_n\}\) would require a complete set of \(O(n^2)\) pairwise comparisons, our objective is to obtain a total ranking \(\sigma\) that satisfies

\[
\forall (I_i, I_j) \in \mathcal{P},\, I_i \succ I_j \implies \sigma(I_i) < \sigma(I_j).
\]

To achieve this efficiently, we employ MergeSort, a stable comparison-based algorithm with time complexity \(O(n \log n)\). MergeSort recursively divides the dataset into smaller subsets, sorts each subset based on the pairwise comparisons, and merges the sorted subsets to produce the final ordering, thereby ensuring that the resulting ranking is consistent and non-cyclic.

\begin{figure}
    \centering
    \includegraphics[width=0.45\textwidth]{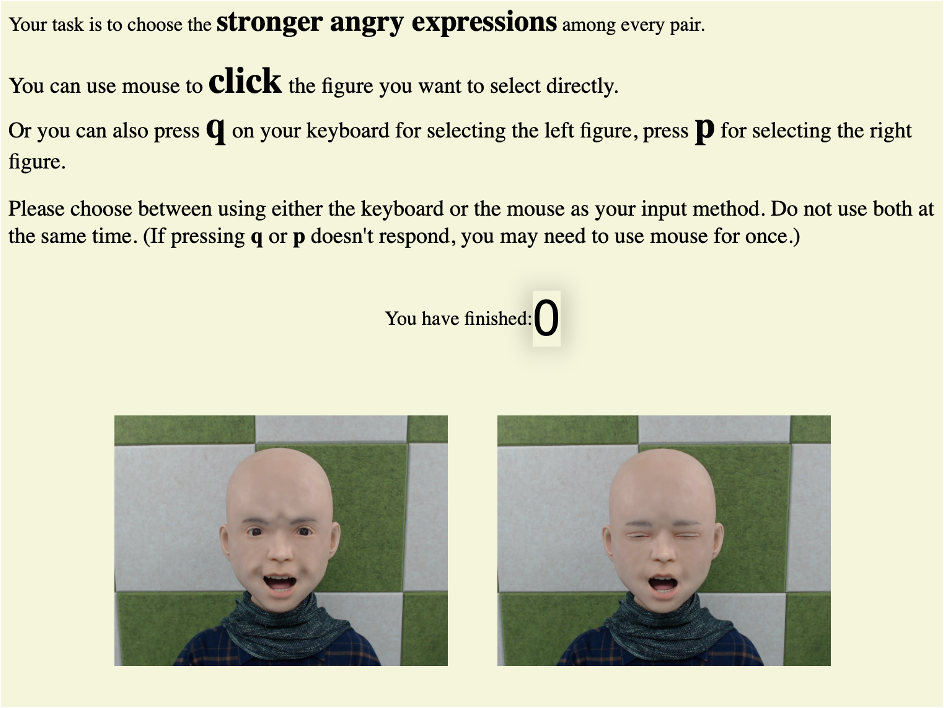}
    \caption{Screenshot of the online annotation platform for pairwise comparison of robot-generated static facial expressions.}
    \label{fig:pairannotation}
\end{figure}

An online platform is developed (Fig. \ref{fig:pairannotation}) to collect pairwise comparisons, where participants are instructed to select the stronger robot facial expression from each pair. We recruit six participants—five Kyoto University students and one research assistant at RIKEN—each of whom annotates all six expressions. Specifically, for each participant and for each expression, we first shuffle the entire dataset and then follow the divide-and-conquer strategy of MergeSort to generate the pairwise comparison trials. Further details on the ranking and pairwise comparison methodology are described above.

To assess the consistency of these rankings with the original pairwise data, we perform a Kendall's Tau evaluation~\cite{kendall1938new}, which yields a correlation coefficient of 100\%. This result indicates perfect agreement. Kendall's Tau is defined as
\[
\tau = \frac{\mathbb{C} - \mathbb{D}}{\mathbb{N}}
\]
where \(\mathbb{C}\) denotes the number of concordant pairs, \(\mathbb{D}\) denotes the number of discordant pairs, and \(\mathbb{N}\) is the total number of pairs. The experiment code is publicly available at our GitHub repository\footnote{\url{https://github.com/KUCognitiveInformaticsLab/PairwiseExpressionAnnotator}}.

\subsection{Human Preference Modeling}

\begin{figure}[htpb]
  \centering
  \includegraphics[width=0.48\textwidth]{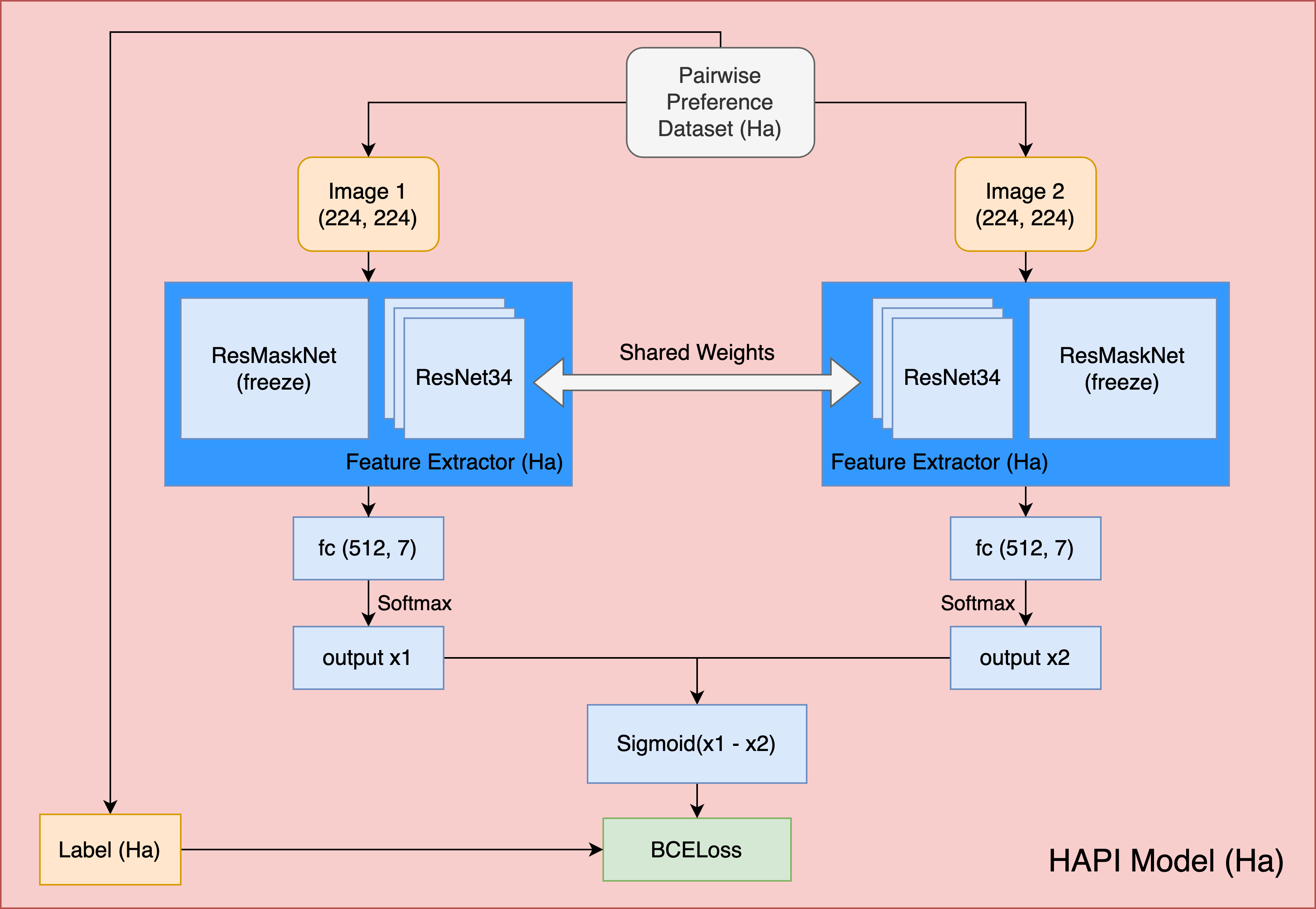}
  \caption{Architecture of the HAPI model for pairwise preference learning of the Happy expression. The model extracts features using ResMaskNet and ResNet34 in parallel, applies shared weights, and computes preference scores using a binary cross-entropy loss.}
  \label{fig:hapimodel}
\end{figure}

Our objective is to model human preferences through comparative judgments. Unlike traditional classification tasks, this problem presents unique challenges in both learning and evaluation due to its inherently subjective nature. To address these challenges, we propose a ResNet-based Siamese RankNet architecture, termed Human Affective Pairwise Impressions (HAPI) model (Figure~\ref{fig:hapimodel}).

The HAPI employs a Siamese network structure with shared weights to process image pairs from a pairwise preference dataset. For a pair of image to be compared, each image is passed through a feature extractor that combines ResMaskNet and ResNet34 in parallel. ResMaskNet is with the residual masking approach originally introduced in~\cite{pham2021facial}, which was trained on the FER2013 dataset for classifying human basic facial expressions. Then, we freeze the ResMaskNet part to retain their pre-trained facial feature capabilities, while keeping the standard ResNet34 layers trainable so they can adapt to preference-based ranking.

Following feature extraction, each processed image is passed through a fully connected (fc) layer (512, 7) with a Softmax activation, generating an output distribution over seven emotional dimensions. The preference score is then computed as the sigmoid difference between the outputs of the two images, reflecting the relative ranking. Finally, Binary Cross-Entropy (BCE) Loss is applied to align the model's predictions of the pairwise preference with human-annotated preferences.

\subsubsection{Baseline}

To benchmark the performance of the HAPI Model, we employ a state-of-the-art pre-trained model introduced in~\cite{pham2021facial}. This model was originally trained on the FER2013 dataset, which comprises 28,709 images for training, 3,859 images for validation, and 3,589 images for testing. The FER2013 dataset was curated using the Google Image Search API, and each image is annotated with one of seven discrete emotion categories: anger, disgust, fear, happiness, sadness, surprise, or neutral.

\subsubsection{Binary Cross-Entropy Loss}

Binary Cross-Entropy (BCE) loss is a standard loss function used for binary classification tasks, quantifying the dissimilarity between the predicted probability distribution and the ground truth labels. For an individual sample, the BCE loss is defined as:

\begin{equation}
    \mathcal{L}_{\text{BCE}}(y, \hat{y}) = - \left[ y \log(\hat{y}) + (1 - y) \log(1 - \hat{y}) \right]
\end{equation}

where:
\begin{itemize}
    \item $y \in \{0, 1\}$ is the ground truth label.
    \item $\hat{y} \in [0, 1]$ is the predicted probability of the positive class.
    \item $\log(\cdot)$ represents the natural logarithm.
\end{itemize}

For a dataset comprising \(N\) samples, the overall BCE loss is computed as the mean loss across all samples:
\begin{equation}
\mathcal{L}_{\text{BCE}} = \frac{1}{N} \sum_{i=1}^{N} - \left[ y_i \log(\hat{y}_i) + (1 - y_i) \log(1 - \hat{y}_i) \right]
\end{equation}

\subsection{Model Fine-Tuning}

\subsubsection{Implementation Details}

All training is conducted on dual NVIDIA A6000 GPUs using PyTorch 1.13.0. We adopt an initial learning rate of 0.005, a weight decay of \(1 \times 10^{-5}\), and a momentum of 0.9. These hyperparameters are selected based on preliminary tuning to balance convergence speed with robust generalization. Specifically, we train six models, each corresponding to one of the six basic expressions (anger, disgust, fear, happiness, sadness, and surprise), ensuring that the learned representations effectively capture the unique characteristics of each expression category.

\subsubsection{Dataset Preparation and Cross-Validation}

We utilize the set of robot facial expressions defined in Section~\ref{sec:hpd}. From this annotated dataset, 20\% of the images are randomly selected to form a validation subset, while the remaining 80\% constitute the training subset. Within each subset, pairwise combinations are generated and labeled to indicate the preferred expression. Given the modest dataset size and sensitivity to test set selection, we adopt a 5-fold cross-validation strategy to mitigate selection bias and ensure robust evaluation.

\subsubsection{Image Preprocessing}

A standardized preprocessing pipeline is applied to each image. First, we employ the face detection algorithm from the Py-Feat toolkit to identify and crop the facial region (facebox). Because the camera setup remains fixed during data collection, facebox coordinates are calculated once and then reused to expedite processing across the dataset. Each cropped image is resized to \(224 \times 224\) pixels, converted to a tensor, and normalized with a mean of \([0.5, 0.5, 0.5]\) and a standard deviation of \([0.5, 0.5, 0.5]\). These steps ensure consistency in input across all images used for model training.

\subsection{Pairwise Preference Performance: HAPI vs. Baseline}

\begin{figure}[htpb]
  \centering
  \includegraphics[width=0.48\textwidth]{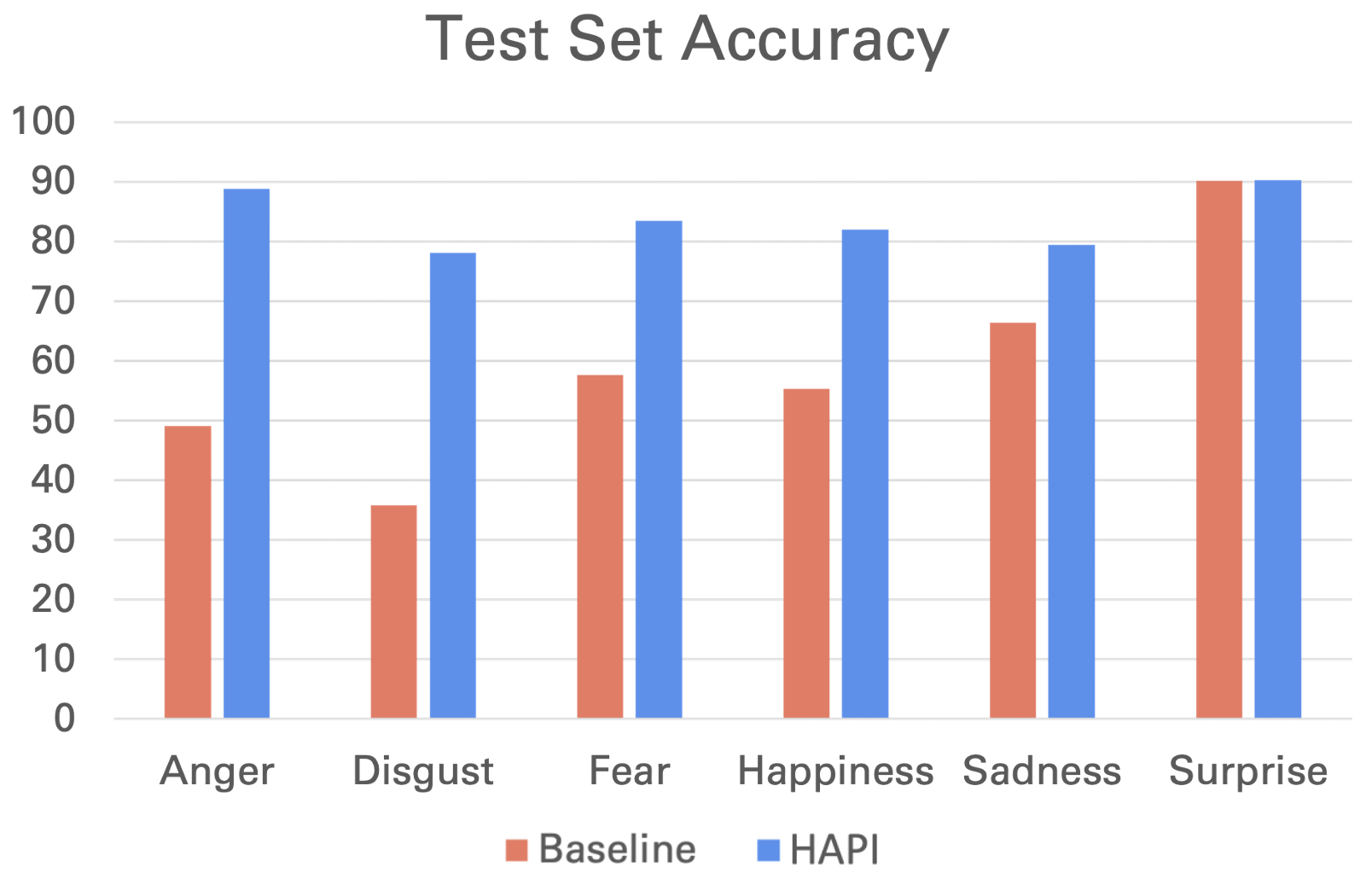}
  \caption{Mean inference accuracy over 5-fold cross-validation for the Baseline and HAPI models across six emotion categories. HAPI consistently outperforms the Baseline, demonstrating improved alignment with human-perceived emotional expressions.}
  \label{fig:modelacc}
\end{figure}

Figure~\ref{fig:modelacc} shows the mean inference accuracy over 5-fold cross-validation for six emotion categories: anger, disgust, fear, happiness, sadness, and surprise. We compare the baseline and HAPI models, observing that HAPI consistently outperforms the baseline for all expressions—with the most pronounced gains in anger and disgust. Notably, HAPI achieves an accuracy above 80\% for most emotion categories, underscoring its enhanced alignment with human-perceived nuances of emotional facial expressions. In contrast, the baseline model struggles, particularly for expressions like disgust. These findings suggest that incorporating human preference data during training substantially improves the model's ability to recognize and refine robotic facial expressions according to human judgments.

\subsection{Robot Expression Generation Experiment}\label{sec:reg}

We further examine the practical value of our framework in an expression generation experiment on the Nikola android robot. Following \cite{yang2022BORFEO}, we employ a Bayesian Optimization strategy guided by both the baseline model and the proposed HAPI. In addition, a method based on Ekman's Facial Action Coding System (FACS)~\cite{ekman1978facial} is incorporated as an expert-designed benchmark.

\begin{figure}[htpb]
  \centering
  \includegraphics[width=0.45\textwidth]{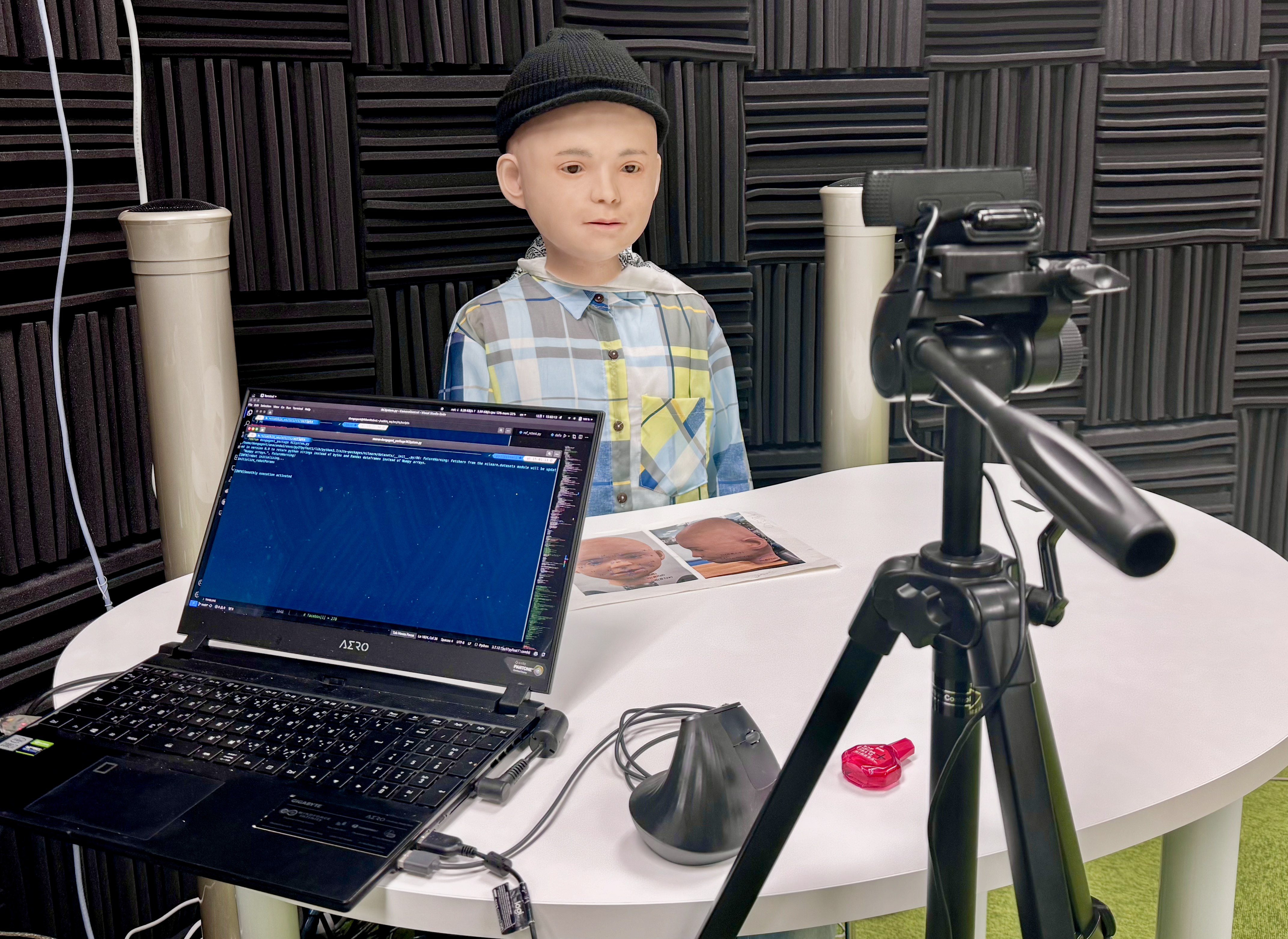}
  \caption{Expression generation experimental setup.}
  \label{fig:expgenset}
\end{figure}

Figure~\ref{fig:expgenset} illustrates the experimental setup. For robot control and camera manipulation, a notebook computer with an Intel Core i7-10870H CPU (2.20 GHz) and an NVIDIA GeForce RTX 3060 laptop GPU (GDDR6 6GB) running Ubuntu 20.04 is used. High-quality RGB images are captured using a Logicool C922 Pro Stream webcam (HD1080P) to ensure consistent and reliable data collection for robot facial expressions. The robot platform, Nikola, operates on ROS (Robot Operating System) Noetic Ninjemys — a system released on May 23, 2020 and fully compatible with Ubuntu 20.04.

Each generation process—targeting anger, disgust, fear, happiness, sadness, or surprise—is iterated 300 times, requiring approximately 20 minutes per session for both baseline and HAPI. All imaging parameters (camera position, exposure, brightness, white balance, and zoom) were fixed to ensure variations in expressions were driven solely by model performance.

\begin{figure}[htpb] 
    \centering 
    \includegraphics[width=0.45\textwidth]{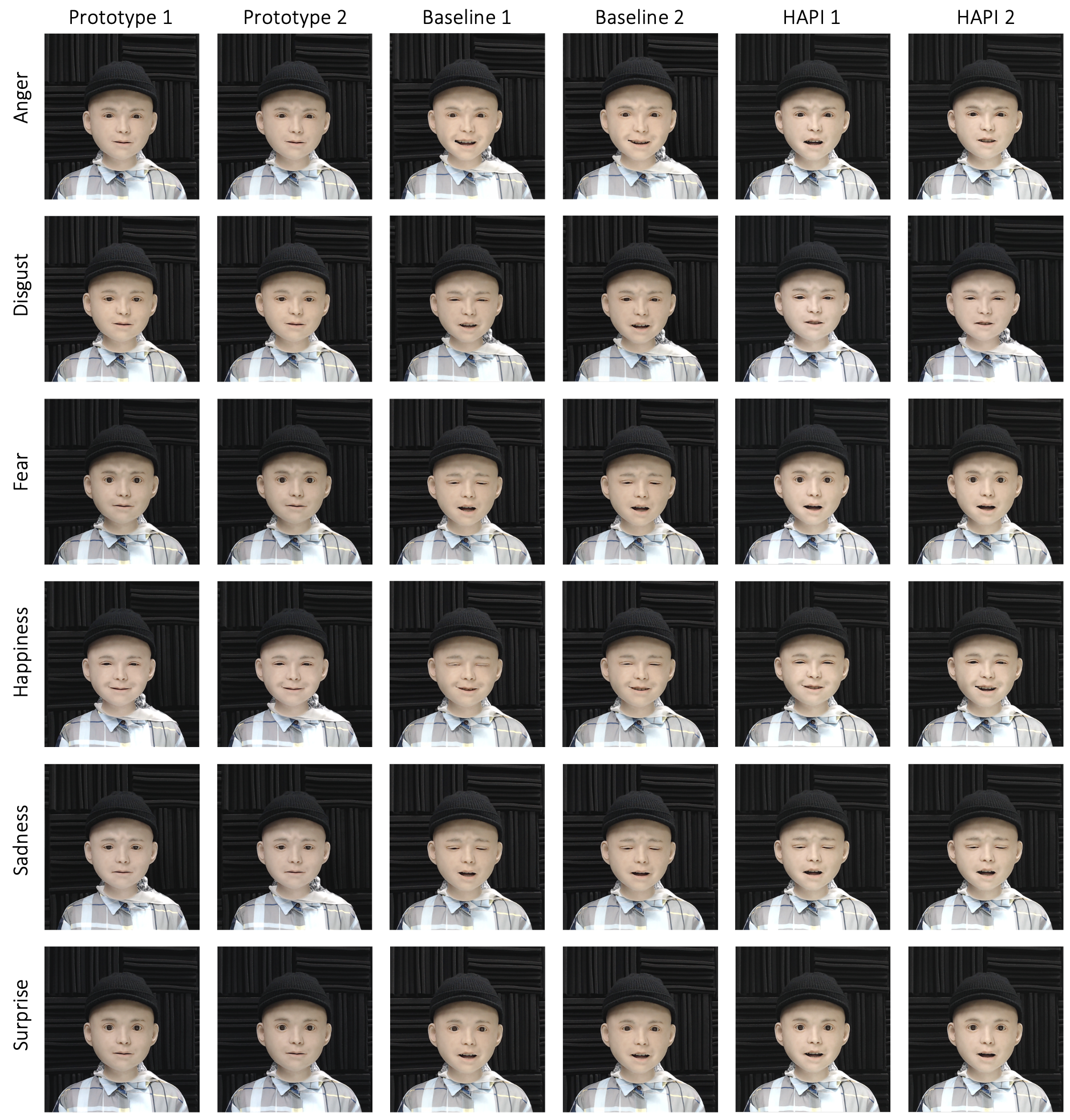} 
    \caption{Robot expression generation results for three methods across six emotion categories.} \label{fig:stim} 
\end{figure}

Figure~\ref{fig:stim} displays representative expressions produced by the three methods—prototype, baseline, and HAPI—across the six emotion categories. For each category, the variations within each method represent subtle differences of the same target expression: prototype and HAPI samples are selected from multiple model-generated candidates using model-assigned preference scores, whereas baseline variations are manually refined by experts. The prototype provides a basic approximation, the baseline introduces moderate enhancements, and HAPI yields the most refined, human-aligned expressions. These results highlight the effectiveness of incorporating human preference data, enabling the generation of robotic facial expressions with finer perceptual granularity.

\section{Crowd Sourcing Validation and Result}

\subsection{Validation Procedure}

We recruit 33 participants (16 male, 16 female, and 1 undisclosed) via the Yahoo crowd-sourcing platform. Each participant evaluates 36 static images covering six emotion categories. For each category, the final stimulus set comprises six images: two prototype-generated expressions (with one repeated), two baseline-generated expressions (top two selected), and two HAPI-generated expressions (top two selected). Participants rate the perceived intensity of all six emotions on a 7-point Likert scale (1 = no perception, 7 = very strong perception). Stimuli (see Figure~\ref{fig:stim}) are randomized to minimize potential order effects. All participants are compensated in accordance with their effort.

\subsection{Statistical Analysis}

We conduct a two-way repeated-measures ANOVA with RFE method (Prototype, Baseline, HAPI) and Emotion Category (Anger, Disgust, Fear, Happiness, Sadness, Surprise) as independent variables. The main effect of Emotion Category is significant, $F(5, 160) = 33.497$, $p < .001$, $\eta_p^2 = 0.511$, and the main effect of RFE method is also significant, $F(2, 64) = 40.519$, $p < .001$, $\eta_p^2 = 0.559$. Moreover, a significant RFE method × Emotion Category interaction emerges $F(10, 320) = 5.229$, $p < .001$, $\eta_p^2 = 0.140$. These findings suggest that both the choice of expression-generation method and the specific emotion category significantly influence perceived intensity, emphasizing the importance of tailoring robotic facial expressions to human affective responses.

\subsection{Post Hoc Comparisons: HAPI vs. Baseline and Prototype}

Pairwise comparisons are performed using Tukey's method to examine simple main effects of the RFE method within each emotion category. Table \ref{tab:posthoc} presents the associated post hoc results.

\begin{table}[htpb]
    \centering
    \caption{Mean difference between method on various emotions}
    \label{tab:posthoc}
    \begin{threeparttable}
    \begin{tabular}{llll}
        \toprule[1.5pt]
        \textbf{Emotion} & \textbf{HAPI vs. B} & \textbf{HAPI vs. P} & \textbf{B vs. P} \\
        \midrule
        \textbf{Anger}    & $0.03702$ \checkmark & $0.06569$ \checkmark  & $0.02866$ \\
        \textbf{Disgust}  & $0.02645$ & $0.09174$ \checkmark & $0.06529$ \checkmark \\
        \textbf{Fear}     & $0.00188$ & $0.01104$ & $0.00917$ \\
        \textbf{Happiness}& $0.08330$ \checkmark & $0.06237$ \checkmark & $-0.02092$ \\
        \textbf{Sadness}  & $-0.01392$ & $0.05489$ \checkmark &  $0.06881$ \checkmark\\
        \textbf{Surprise} & $0.06501$ \checkmark & $0.10819$ \checkmark & $0.04318$ \checkmark\\
        \bottomrule[1.25pt]
    \end{tabular}
    \begin{tablenotes}
        \item \checkmark indicates statistical significance at $p < 0.05$.  
        \item \textbf{B}: Baseline. \textbf{P}: Prototype.
    \end{tablenotes}
    \end{threeparttable}
\end{table}

\begin{itemize}
    \item \textbf{Anger}: HAPI is rated significantly higher than Prototype (\( p < .0001 \)) and Baseline (\( p < .05 \)).  
    \item \textbf{Disgust}: HAPI is rated significantly higher than Prototype (\( p < .0001 \)), and Baseline is also rated significantly higher than Prototype (\( p < .0001 \)).
    \item \textbf{Fear}: No significant differences are observed among the three methods.
    \item \textbf{Happiness}: HAPI is rated significantly higher than both Baseline and Prototype (\( p < .0001 \)).  
    \item \textbf{Sadness}: Both Baseline (\( p < .0001 \)) and HAPI (\( p < 0.001 \)) are rated significantly higher than Prototype. 
    \item \textbf{Surprise}: HAPI is rated significantly higher than both Baseline and Prototype (\( p < .0001 \)), and Baseline is also rated significantly higher than Prototype (\( p < 0.01 \)).  
\end{itemize}

\begin{figure}[htpb]
  \centering
  \includegraphics[width=0.48\textwidth]{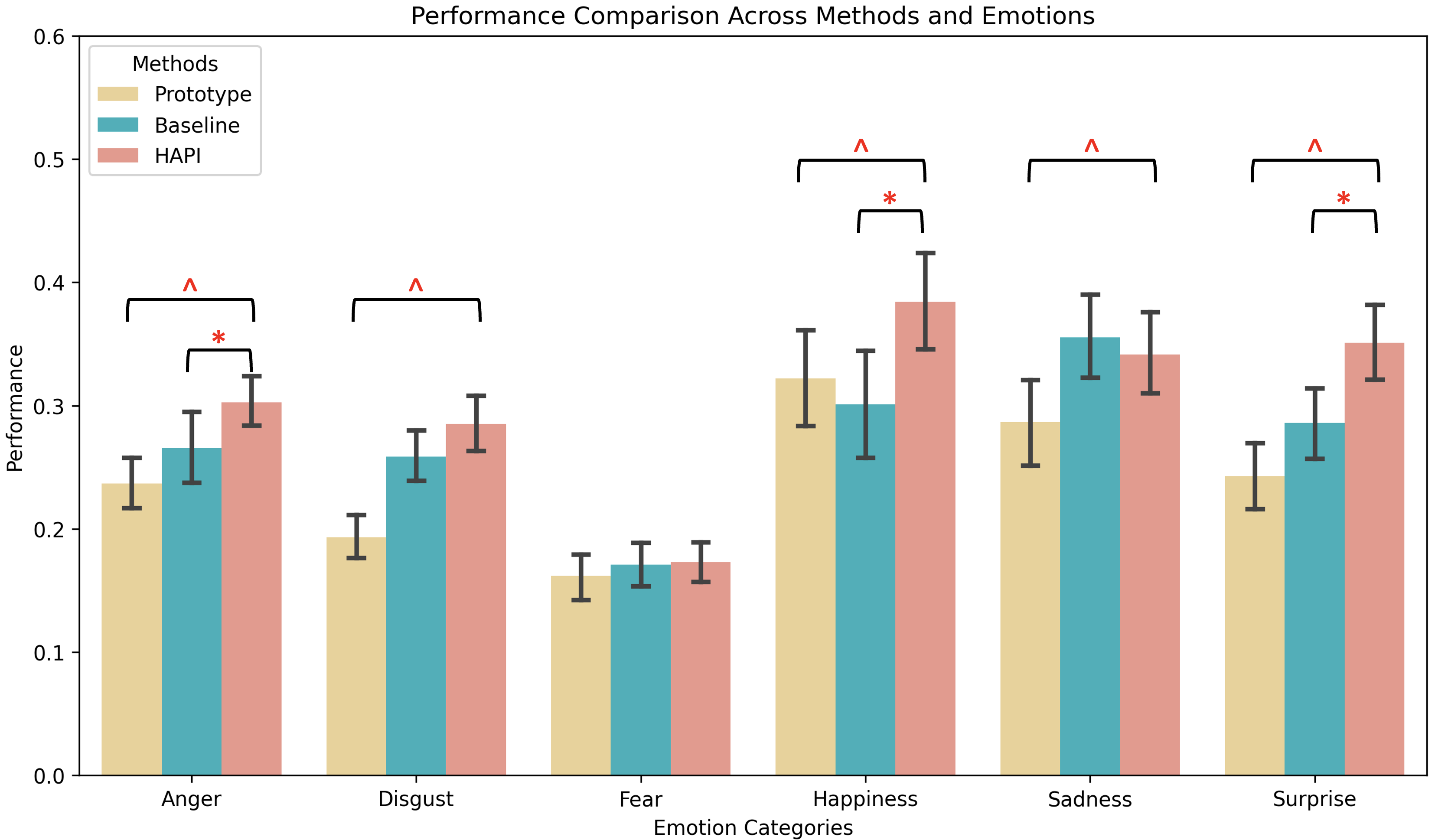}
  \caption{Performance comparison of robotic facial expression generation methods across six emotion categories.  Error bars represent 95\% confidence intervals. \textbf{(*)} indicates a statistically significant difference between \textbf{HAPI and Baseline} (\(p < 0.05\)), while \textbf{(\^{})} denotes a statistically significant difference between \textbf{HAPI and Prototype} (\(p < 0.05\)), based on Tukey-adjusted pairwise comparisons.}
  \label{fig:res}
\end{figure}

\subsection{Result}
Figure~\ref{fig:res} presents the comparative performance of the three robotic facial expression generation methods, along with the corresponding statistical analyses. Overall, HAPI significantly enhances the perceived intensity of robotic facial expressions, particularly for Anger, Happiness, and Surprise, where it consistently outperforms both Baseline and Prototype. For Disgust and Sadness, HAPI also demonstrates significant improvements over Prototype, though its advantage over Baseline is comparatively modest. In the case of Fear, none of the methods exhibits a significant advantage.

\section{Discussion and Future Work}

Our findings demonstrate that the HAPI framework effectively aligns robotic facial expressions with nuanced human preferences, achieving particularly strong outcomes for Anger, Happiness, and Surprise (Figure~\ref{fig:res}). By integrating human preference learning through a Siamese RankNet architecture and leveraging pairwise comparison data, HAPI offers a robust and adaptable approach to generating more realistic and socially resonant expressions.

Despite these advances, there are several areas for improvement that also represent opportunities for future development. First, the limited performance in conveying Fear and Sadness highlights the importance of addressing hardware constraints—for example, repairing or upgrading the lip corner puller servo to unlock a fuller range of emotional expressiveness. Second, ensuring culturally diverse training data is vital for capturing global variations in facial expressions; expanding beyond FER2013 would help refine performance for emotions like Fear and Disgust that can vary significantly across cultural contexts.

From a deployment perspective, HAPI's modular design already facilitates integration with ROS-based and other servo-controlled systems, making rapid customization feasible for various real-world applications. Future steps include enhancing the pairwise comparison mechanism—potentially incorporating reinforcement learning from human feedback (RLHF)—to further refine subtle emotional cues and improve expression authenticity. Additional work on hardware adjustments, data diversity, and streamlined system integration will solidify HAPI's generalizability and pave the way for even more compelling human–robot interactions in diverse environments.

\section{Conclusion}

In summary, we introduce a learning-to-rank framework that integrates human perceptual insights into robotic facial expression generation, bridging the gap between conventional quantitative models and the nuances of human affective preference. By conducting pairwise comparison annotations, we collect a robust set of human preference data and then leverage a Siamese RankNet architecture (HAPI) to effectively model these subjective preferences. The HAPI framework represents a significant advancement in robotic facial expression generation, integrating human preference learning and nuanced ranking mechanisms to enhance expressiveness. Experimental results (Figure~\ref{fig:res}) demonstrate its superiority in generating Anger, Happiness, and Surprise, along with measurable improvements in Disgust and Sadness compared to Baseline and Prototype methods. Moreover, HAPI's robust and adaptable design—including seamless integration with ROS-based systems and other servo-controlled platforms—positions it as a versatile tool for real-world applications. By laying a strong foundation for the next generation of expressive robotic systems, HAPI advances human–robot interaction through more natural and emotionally resonant facial expressions.

Despite these challenges, HAPI's robust design and adaptability—including seamless integration with ROS-based systems and other servo-controlled platforms—position it as a versatile tool for real-world applications. Future work will focus on hardware improvements, dataset diversification, and refinement of learning mechanisms to enhance both realism and cultural universality. By addressing these areas, HAPI lays a strong foundation for the next generation of expressive robotic systems, advancing human-robot interaction through more natural and emotionally resonant facial expressions.

\section*{ACKNOWLEDGMENT}

This work was supported in part by JST SPRING, Grant Number JPMJSP2110.

\bibliographystyle{IEEEtran} 
\bibliography{./IEEEabrv,./IEEEexample}

\end{document}